\documentclass[]{llncs}
\usepackage{epsf}

\usepackage{url}  
\usepackage{graphicx}  

\usepackage{filecontents}
\usepackage{wrapfig}

\usepackage[utf8]{inputenc} 
\usepackage[T1]{fontenc}    
\usepackage[hidelinks]{hyperref}       
\usepackage{lmodern}
\usepackage{amsfonts}       
\usepackage{amsmath}
\usepackage{amssymb}
\usepackage{nicefrac}       
\usepackage{microtype}      
\usepackage[ruled]{algorithm2e}
\usepackage[english]{babel}
\usepackage{color}
\usepackage{latexsym}
\usepackage{multirow}
\usepackage{subcaption}
\usepackage{tikz}
\usepackage{pgfplots}

\usepackage{pgfplotstable}
\usepackage{filecontents}
\pgfplotsset{compat=1.13}
\usetikzlibrary{patterns}
\usepgfplotslibrary{colormaps} 

\definecolor{awesome}{rgb}{1.0, 0.13, 0.32}
\definecolor{magenta_mtplotlib}{rgb}{1.0, 0.0, 1.0}
\definecolor{green_mtplotlib}{rgb}{0.0, 0.5019607843137255, 0.0}
\definecolor{blue_mtplotlib}{rgb}{0.0, 0.0, 1.0}

\title{Learning compositionally through attentive guidance}

\author{
Dieuwke Hupkes$^1$ and Anand Kumar Singh$^2$ and Kris Korrel$^2$ and German Kruszewski$^3$ and Elia Bruni$^1$\\
}

\institute{
$^1$ ILLC, University of Amsterdam\\
$^2$ University of Amsterdam\\
$^3$ Facebook AI Research\\
~\\
\{d.hupkes, e.bruni\}@uva.nl, \{anand.singh, kristan.korrel\}@student.uva.nl, germank@fb.com
}

\begin{document}

\maketitle

\begin{abstract}
While neural network models have been successfully applied to domains that require substantial generalisation skills, recent studies have implied that they struggle when solving the task they are trained on requires inferring its underlying compositional structure.
In this paper, we introduce \textit{Attentive Guidance}, a mechanism to direct a sequence to sequence model equipped with attention to find more compositional solutions.
We test it on two tasks, devised precisely to assess the compositional capabilities of neural models, and we show that vanilla sequence to sequence models with attention overfit the training distribution, while the guided versions come up with compositional solutions that fit the training and testing distributions almost equally well.
Moreover, the learned solutions generalise even in cases where the training and testing distributions strongly diverge.
In this way, we demonstrate that sequence to sequence models are capable of finding compositional solutions without requiring extra components.
These results helps to disentangle the causes for the lack of systematic compositionality in neural networks, which can in turn fuel future work. 
\end{abstract}

\section{Introduction}

In the past decade, neural network research has made astonishing progress on many tasks that require significant generalisation abilities, such as machine translation, visual reasoning and playing complex games such as Go \cite{sutskever2014sequence,silver2016mastering,johnson2017clevr}.
However, despite these empirical successes on human tasks, the form of generalisation such models exhibit diverges strongly from human type generalisation, which becomes visible when attacking systems with adversarial examples \cite{bender2017proceedings} or testing them on data with a different distribution than the training data \cite{lake2017systematic}.

One key problem appears to be that neural networks trained with our current learning paradigms do not find solutions exhibit systematic compositionality \cite{livska2018memorize,lake2017systematic}, an element which has shown to be crucial for systems to behave in a more human-like manner \cite{lake2015human}.
Neural networks seem to rely strongly on pattern matching and memorisation, but humans -- on the other hand -- are known to have a strong preference for systematic, compositional solutions \cite{schulz2016probing}.
To give an example: If a human knows how to walk from their hotel (site \texttt{000}) to the conference site (\texttt{110}) via road \texttt{t1} and to reach a nice restaurant (\texttt{011}) from there via road \texttt{t2}, they won't have troubles to go from their hotel to the restaurant by composing these two routes, even if they are otherwise unfamiliar in the city (we will revisit this example later in the first of our experiments).

The lack of systematic compositionality in the solutions found by neural networks does not only result in data-hungry models that require large amounts of training data, but also severely limits their applicability in domains that require exactly this type of generalisation, such as language.
An increasing amount of hybrid approaches that explicitly build in compositional components in connectionist models \cite{reed2015neural,socher2010learning,kuncoro2018lstms}, but little is known about inducing compositional solutions in purely connectionist models, such as LSTMs \cite{hochreiter1997long} and GRUs \cite{chung2014empirical}.
Understanding how to make compositional solutions emerge without losing the attractive properties of vanilla recurrent models provides an advantage over hand-engineering compositional structures, as the latter does not easily generalise to different domains and might require very large amounts of data.

In this paper, we hypothesise that this lack of compositionality is due to recurrent models' disability to distinguish salient patterns from spurious ones.
For instance, from a linguistics perspective, in the sentence \textit{the man with the hat walks in the park}, \textit{walks in the park} may be such a reusable subsequence, whereas \textit{hat walks in} is not, regardless of how often it may have occured in a corpus.

We explore a novel approach that addresses exactly this problem, by providing additional information to the attention component of a vanilla seq2seq \cite{sutskever2014sequence,cho2014learning} model during training.
This extra information, which we call \textit{Attentive Guidance} (\textbf{AG}), guides the model to compositionally attend to the input while generating its output, and with this provides a learning bias that guides the model to converge to more compositional solutions.

We test AG on two different datasets that were both specifically designed to test the compositional generalisation abilities of neural networks: the lookup table task \cite{livska2018memorize} and the symbol rewriting task \cite{weber2018fine}.
On both datasets, using AG consistently and robustly changes the types of solutions found to be more compositional.

\section{Related work}

The ability to learn compositionally by combining small components into larger parts is considered one of the hallmarks of human intelligence \cite[]{szabo2010compositionality,lake2015human}.
Compositionality and systematicity have long since played a prominent role in research about language \cite{chomsky1956three,fodor1988connectionism}, where it explains that we are able to comprehend expressions that we have never seen or heard before but are composed of parts that we already know, but it has also been proposed as a core aspect of understanding visual concepts \cite{lake2015human,schulz2016probing} and even of our general motor skills \cite{flash2005motor}.

Given the integral importance of compositional learning, designing machine learning models that are capable of discovering and exploiting compositional explanations for data has received a lot of interest in different communities.
One important line of work focuses on combining separate custom modules in a more complex architecture \cite{andreas2016modular,johnson2017clevr}.
However, the modules in this case are typically fixed and not learned, while the whole system is trained on program traces, and thus it receives a strong supervision on how to decompose the tasks.
An approach to address this is to supervise (neural network) modules on explicitly defined data transformations and let a controller issue the desired sequence of operations \cite{reed2015neural,kurach2015neural,neelakantan2015neural}, that can potentially be recursive \cite{cai2017making}.
These models can learn to perform well on complex tasks, but require quite heavy supervision of both the modules and the controller, and are constraint in terms of the number of available data transformations and how they can be combined.
Another approach that is more akin to the one we take is that of \cite{hudson2018compositional}, where, in the context of visual reasoning, they propose a new differentiable architecture specifically designed to facilitate reasoning. 

While some of these newly introduced models are promising steps towards finding more generally applicable solutions for compositional learning, they still lack the generality of popular recurrent sequence models such as LSTMs \cite{hochreiter1997long} and GRUs \cite{chung2014empirical}.
The latter, however, have been shown to struggle when it comes to learning compositional rules that allow for generalisation outside their train set distribution. 
For example, \cite{lake2017systematic} and \cite{livska2018memorize} show that when the training and testing distributions are manipulated such that generalisation requires finding such compositional structures, vanilla LSTMs do not succeed at generalisation to this testing space.

In this paper, we explore whether supervising the model's attention can help it generalise in more systematic ways. 
This technique has been tried before in the context of VQA \cite{qiao2017exploring,gan2017vqs} and Machine Translation \cite{mi2016supervised}.
Differently from them, here we use it to test our specific hypothesis.

\section{Attentive Guidance}\label{sec:att_guidance}

It is well known that a finite set of input-output pairs can be described in many different ways \cite{angluin1983inductive,gold1967language}.
In the space of all potential solutions, one extreme is to simply memorise all pairs, which does not offer any means to extrapolate to new inputs; on the other end of the spectrum we find solutions that abide by the principle of minimal description length, which provides much stronger generalisation capacity \cite{hutter2004universal,rissanen1978mdl}. 
Vanilla seq2seq models can in theory represent many different solutions on this spectrum, but -- as it turns out -- are very unlikely to converge to the type of compositional solutions that we desire when they are trained with our current learning paradigms.

We hypothesise that this is -- in part -- due to the fact that, within the current learning paradigms, very little information is provided regarding the type of solutions that we would like a model to find.
As a consequence, models are unable to distinguish salient from non-salient patterns and end up memorising patterns that may seem reasonable given the training data, but are not when considering the underlying system. 
For instance, a model might memorise the previously mentioned composed path from hotel to conference to restaurant as a whole and may then not be able to go to the restaurant from the conference site if the latter was reached from a gym rather than the hotel.

In this paper, we investigate if we can use the attentive component of a seq2seq model to explicitly tell the model what the salient subsequences of the input are and how they should be attended in a compositional way and as such bias models to find more compositional solutions.


Concretely, we transmit this bias to a network by adding an extra loss term to the objective function used to train a model, which represents the difference between the model's computed attention and the target attention pattern:

\[\mathcal{L}_{\text{AG}} = \frac{1}{T}\left( \sum_{t=1}^T{\sum_{i=1}^N} -a_{i,t} \log\hat{a}_{i,t}\right), \]
\noindent where $T$ is the length of the target output sequence, $N$ is the length of the input sequence, $\hat{a}_{i,t}$ is the decoder-computed attention of input token $i$ at time $t$ and $a_{i,t}$ is its corresponding attention target.
We weight the original loss of the model and the AG loss with their corresponding weighting factors.

\begin{figure*}[]
    \centering
    \includegraphics[width=0.99\linewidth]{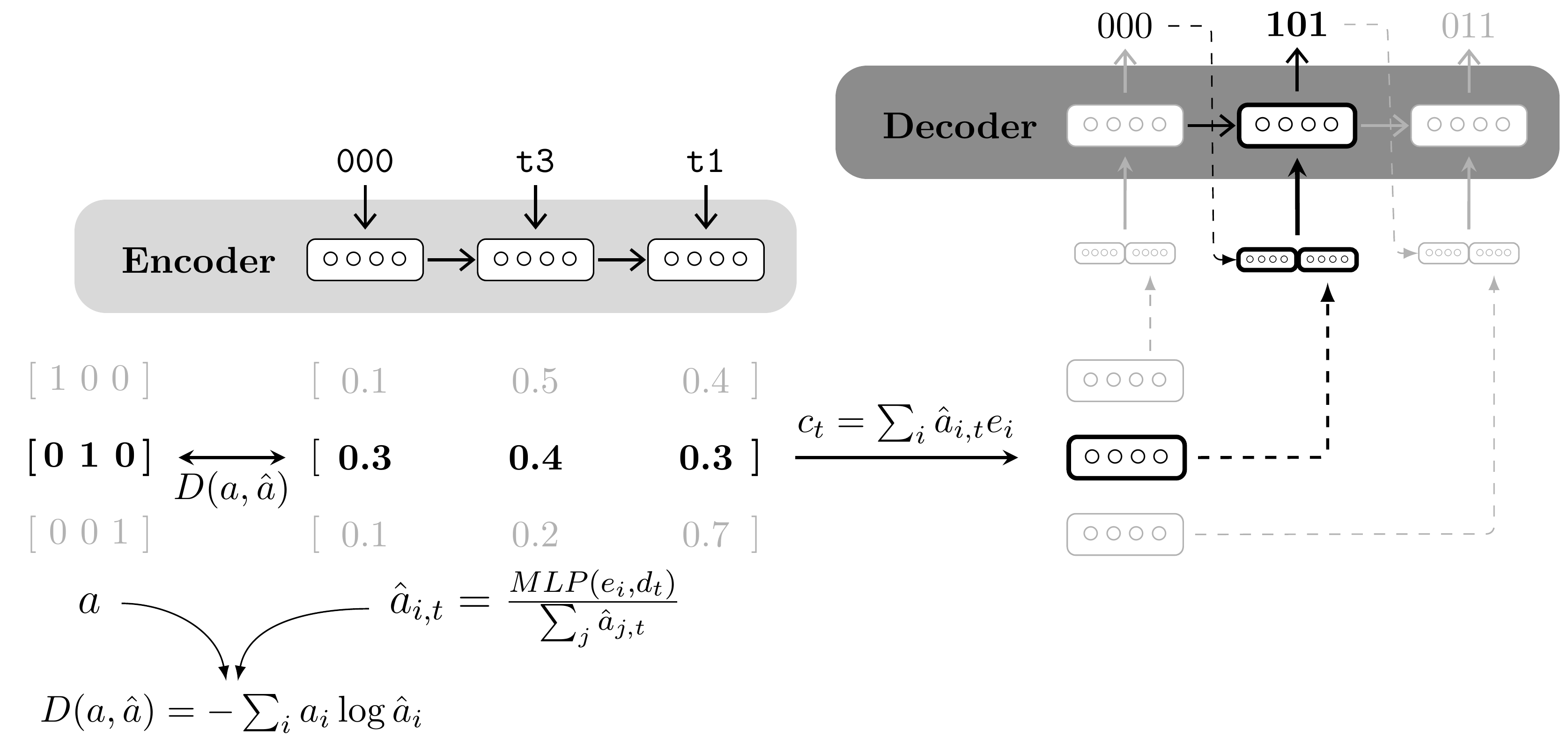}
    \caption{
At each time step $t$, the AG-guided model infers a variable-length alignment weight vector $\hat{a}$ based on the current target state $h_{t}$ and all source states. The Cross Entropy $D$ between $\hat{a}$ and the one-hot AG target $a$ is computed. A context vector $c_{t}$ is calculated as the weighted average, according to $\hat{a}$ over all the encoder states and is then concatenated with the output of the previous step to generate the next prediction. }\label{fig:attentive_guidance} 
\end{figure*}

The target attention pattern specifies how the input should be attended in a compositional way, we thus call the use of this additional loss term AG.
A schematic of attentive guidance is given in Figure \ref{fig:attentive_guidance}.\footnote{If the paper is accepted, we will make our code available.}

\section{Experiments}

To test the effectiveness of AG, we look at two different data sets, both specifically designed to test the compositional generalisation abilities of recurrent models: the \textit{lookup tables} task \cite{livska2018memorize} and the \textit{symbol rewriting} task \cite{weber2018fine}.
We discuss both datasets below.

\subsection{Lookup tables}

\cite{livska2018memorize} proposed the lookup table composition task as a simple setup to test compositional behaviour.
The task consists in computing the meaning of sequences of \textit{lookup tables}.
These lookup tables are bijective mappings from one binary string of length $L$ to another binary string of the same length and are -- together with their input -- presented to a network as sequences of words.
For instance, a potential input sequence could be \texttt{000 t1 t2} (from the hotel take path \texttt{t1} and then path \texttt{t2}); computing its meaning would involve first looking up $\texttt{000 t1}$ (where do I get when I take path \texttt{t1} from the hotel), and then applying \texttt{t2} to its result. \footnote{Note that we use \textit{polish} notation, to facilitate an incremental computation of the outcome.}

Because applying the lookup tables themselves is nothing more than rote memorisation, the difficulty of the task resides solely in inferring the compositional structure from the input-output sequences.
The task performance of the network is thus directly linked to the extent to which the solution a network inferred is compositional, without being conflated by other factors.
\cite{livska2018memorize} show that a vanilla LSTM model converges to a solution that generalises to the test set for only very few initialisations, while the vast majority of trained networks does not exhibit compositional behaviour. 

\begin{figure}
    \centering
    \begin{tabular}{clll}
        & \textbf{\small Input} \vspace{1mm}  & \textbf{\small Target}    & \textbf{\small AG Target}\\
  \textit{Atomic}   & \texttt{000 t1} & \texttt{000 011} & \texttt{0 1}\\
                    & \texttt{001 t2}    & \texttt{001 000} & \texttt{0 1}\\
                    & \ldots &  \ldots\\
  \textit{Composed}\hspace{2mm} & \texttt{001 t2 t1} \hspace{3mm} & \texttt{001 000 011} \hspace{3mm} & \texttt{0 1 2}\\
                    & \texttt{110 t2 t3} & \texttt{110 001 000} & \texttt{0 1 2}\\
                          & \ldots &  \ldots\\
    \end{tabular}
\caption{Examples of 3bit lookup tables and a length-two composition. Note that the order of presentation is following \textit{Polish} notation, allowing the encoder to process the input right away rather than having to wait until the very last input symbol. The AG target is a sequence of indices that represent to which symbols in the input sequence the decoder should attend.}\label{tab:data_lookup}
\end{figure}

\subsubsection{Data} \label{lt_data}
Following \cite{livska2018memorize}, we use 8 randomly generated 3-bit atomic lookup tables and the 64 possible length-two compositions of these atomic tables. 
For each of these 64 compositions, we randomly take out 2 (out of 8) inputs for testing, a condition we call \textit{heldout inputs}.
While \cite{livska2018memorize} consider only this condition, we create two incrementally more difficult conditions: \textit{heldout compositions} and \textit{heldout tables}.
For the heldout compositions condition, we remove 8 randomly chosen compositions from the training set entirely, allowing us to test how the model generalises to combinations of tables that never occurred in the training set.
With the heldout tables condition, we test to what extent the model can generalise to compositions with tables \texttt{t7} and \texttt{t8} that occur in the training set \textit{atomically} (e.g. \texttt{000 t7}) but never in any composition.
The training set consists of remaining atomic tables (excluding \texttt{t7} and \texttt{t8}) and remaining length-two compositions.

\subsubsection{Training signal}
Following \cite{livska2018memorize}, we define the training signal to be the entire sequence of outputs for the input sequence, including the intermediate computation steps. 
To the beginning of the target sequence, we append an additional step which asks to identify the \textit{input} to the sequence of tables.
Some examples of input-output sequences can be found in Figure \ref{tab:data_lookup}.

\subsubsection{AG signal}

We define the AG target as a sequential reading of the input: during the first step, the network should identify the input of the composition (\texttt{000} in Figure~\ref{fig:attentive_guidance}), in the second and third step, the targets corresponds to the first and second table that should be applied (\texttt{t3} and \texttt{t1} in Figure~\ref{fig:attentive_guidance}).  

\subsection{Symbol rewriting}
To demonstrate the effectiveness and generalisability of AG, we test its performance also on a second task, proposed recently by \cite{weber2018fine}, to test to what extent neural networks models can infer linguistic-like structure from a dataset.
The task they propose consists in rewriting a sequence of input symbols $\{x_1, \dots, x_n\}$ to a sequence of output symbols $\{Y_{i,1}, \dots, Y_{i,n}\}$, following a simple context-free grammar.
Every of the 40 input symbols $x_i$ should be rewritten as a sequence of 3 (distinct) symbols from its own output alphabet $Y_i$, where each symbol in this alphabet can take on 16 different values.


\cite{weber2018fine} show that seq2seq models trained on this task perform consistently very well on different examples of the same distribution, but generalise very poorly to sequences with fewer or more input symbols than seen in training.
This indicates that also for this task models rely more on memorising spurious patterns than inferring the underlying pattern, which makes this task very suitable as testbed for AG.

\subsubsection{Data}
For our experiments we use the exact same data as \cite{weber2018fine}, whose training data consist of 100.000 pairs with input lengths within $[5-10]$.
Crucially, there are no repetitions of symbols in the input sequences.
There are four different test sets, that test generalisation capacity in different scenarios.
The examples in the \textit{Standard} test set are drawn from the same distribution as the train data; \textit{Repeat} is a set where the length distribution is kept identical, but repetitions introduced in the input sequences are allowed; \textit{Short} consists of shorter input sequences ($[1,4]$), and \textit{Long} consists of longer sequences ($[11,15]$).
All data sets are non-exhaustive and sampled randomly from all valid input-output pairs.
For model selection, a validation set is created by sampling inputs with length $[3-12]$ (including inputs with repetitions), thus containing examples from all different testing conditions.

\subsubsection{Training signal}
Following \cite{weber2018fine}, we define the training signal on the target output that occurs in the dataset for a particular input sequence.
Note that because the mapping from input to output is not a function (one input is connected with multiple outputs), a grammatically correct output that is different from the target output available in the training set will be (partly) considered a mistake.

\subsubsection{AG signal}

For every output symbol, we define the attention target to be the position of the input word from which it was generated.
That is, for the first three output symbols, the attention target corresponds to the first input symbol, the second three output symbols should attend to the second input symbol, and so on.

\begin{figure}[t]
    \centering
    \begin{tabular}{ll}
        \multicolumn{2}{l}{$\mathcal{L}$: $X=\{A, B\}$,}\\
        \multicolumn{2}{l}{$Y_A = \{a_1, a_2, a_3\}$, $Y_B = \{b_1, b_2, b_3\}$.}\\
        \multicolumn{2}{l}{$a_1 \rightarrow a_{11} | a_{12}$,\; $a_2 \rightarrow a_{21} | a_{22}$,\; $a_3 \rightarrow a_{31} | a_{32}$}\\
        \multicolumn{2}{l}{$\:b_1 \rightarrow b_{11} | b_{12}$,\;\; $b_2 \rightarrow b_{21} | b_{22}$,\;\; $b_3 \rightarrow b_{31} | b_{32}$}\\\\
    \end{tabular}
    \begin{tabular}{lll}
        \textbf{Input} \vspace{1mm} \hspace{2mm} & \textbf{Valid output for $\mathcal{L}$} & \textbf{AG target}\\
        $AAB$ & $a_{21} a_{32} a_{12} a_{11} a_{22} a_{32} b_{13} b_{21} b_{32}$ \hspace{2mm} & \texttt{000111222}
    \end{tabular}
\caption{An example grammar for the symbol rewriting task, together with a valid input-output pair for this grammar.}\label{tab:data}
\end{figure}



\section{Results}
\label{sec:results}

We compare a standard encoder-decoder model (\textit{baseline}) with models trained with AG (\textit{guided}).
The two model types are identical from an architectural perspective, but -- as described before -- the guided model has the additional training objective to minimise the cross-entropy loss between the calculated attention vectors and a provided target attention vectors in its backward pass.
To take apart the learnability of the target attention patterns and their effect on the learned solution, in the analysis section of this paper we evaluate also the impact of replacing the calculated attention vectors in the forward pass of a model with the target attention vectors (\textit{oracle guidance}).
%


\subsection{Lookup Tables}

To understand the robustness of AG over different parameter settings, we run a grid search over multiple model sizes and attention mechanisms.\footnote{We found that GRU models behave better for the lookup table task for both AG and guided models. In our further experiments with lookup tables, we therefore only use GRU units.}
For each condition (\textit{baseline} and \textit{guided}), we search through embedding and hidden layer sizes 
\{16, 32, 64 and 128\} and \{32, 64, 128, 256 and 512\}, respectively.
Furthermore, we test three different alignment models for the attention mechanism:
\textit{dot} where the attention weights are computed by taking a dot-product between the encoder and decoder hidden state; \textit{mlp}, where instead a multilayer perceptron (MLP) is used and \textit{full focus}, a new alignment mechanism where the context vector is used to gate the decoder input.
We experiment with computing the attention before (\textit{pre-rnn} attention) and after (\textit{post-rnn} attention) the recurrency in the decoder (following \cite{bahdanau2015attention} and \cite{luong2015effective}, respectively). 
More details about the different attention mechanisms can be found in the supplementary materials.
We run each configuration 3 times and investigate the development of the accuracy of the resulting models on \textit{heldout inputs}, \textit{heldout tables} and \textit{heldout compositions}. 

\subsubsection{Model size}
\begin{figure}[]
    \centering
    \includegraphics[width=.7\linewidth]{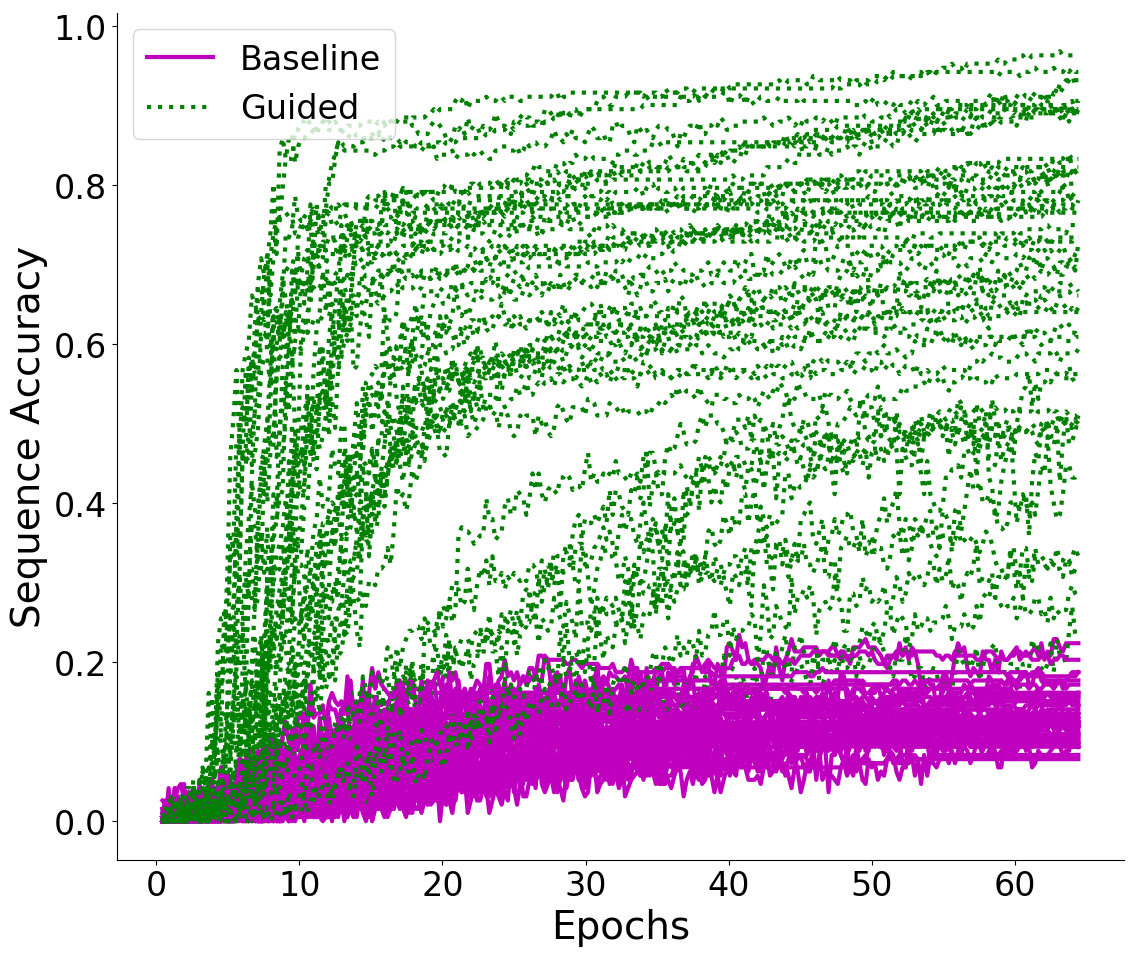}
    \caption{Accuracy on the easiest generalisation condition -- \textit{heldout} tables -- for all models in our grid search. The plot illustrates the accross the board difference between the two models: even the worst models trained with attentive guidance (green dotted lines) generalise better than the best baseline models (magenta lines).}\label{fig:overfitting_accuracy}
\end{figure}

Our grid search confirms the findings of \cite{livska2018memorize} that vanilla recurrent models cannot solve the lookup table task, irrespective of alignment model, place of the attention and network size.
The baseline performance slightly increases with hidden layer size, but never reaches an average accuracy of higher than 30\% across the heldout data, even though all models have a near-perfect performance on the training data.
The guided models, on the other hand, appear to require a certain hidden layer size to generalise well, although even the smallest guided models outperform the larger baseline models.
To illustrate, we depict the development of the accuracy for the \textit{heldout tables} case for all baseline (magenta) and guided (green) models in Figure~\ref{fig:overfitting_accuracy}.

\begin{figure}[h]
    \centering
\includegraphics[width=.7\linewidth]{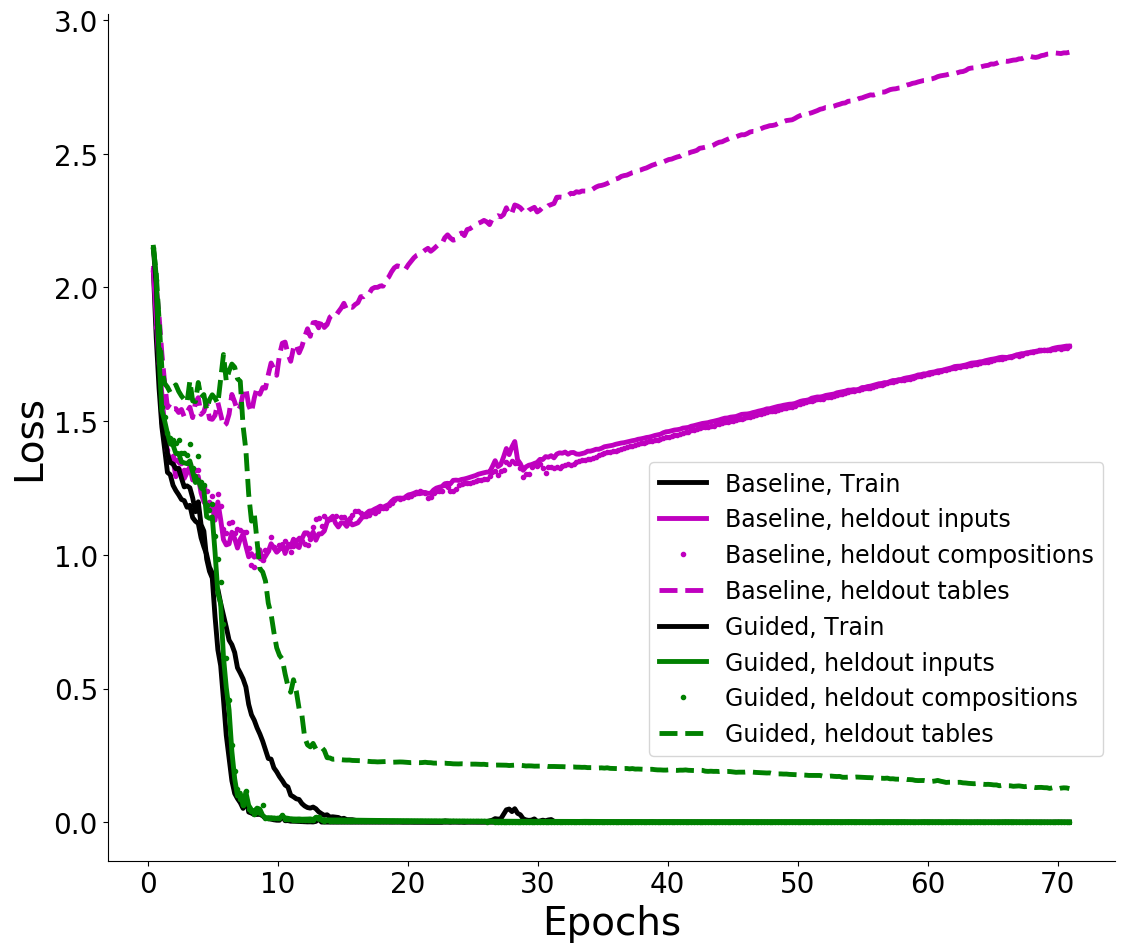}
    \caption{Typical loss development for baseline and guided models on all different test sets. Although both the baseline and models trained with attentive guidance converge on the training set (black lines), the baseline model overfits the training data, as indicated by the increasing loss for the different testsets (magenta lines). The guided models, on the other hand, do not overfit at all: their loss on the test sets only decreases as the training progresses (green lines).}\label{fig:overfitting_loss}
\end{figure}

Furthermore, all baseline models exhibit overfitting on the heldout data, whereas the guided models do not overfit at all: although smaller models do not always reach a high validation accuracy, their validation loss never increases during training.
In Figure~\ref{fig:overfitting_loss}, we plot the development of the loss on all test sets for a typical run of the best baseline and guided configurations.
Importantly, while the training loss converges quickly, the validation loss keeps decreasing steadily for much longer, which can be explained by the influence of the AG loss.

\subsubsection{Attention mechanisms}

The baseline models are indifferent with respect to both the alignment model and the location of the attention.
For the guided models, the \textit{full focus} alignment model combined with pre-rnn attention outperform all other conditions.


\subsubsection{Evaluation of best configurations}

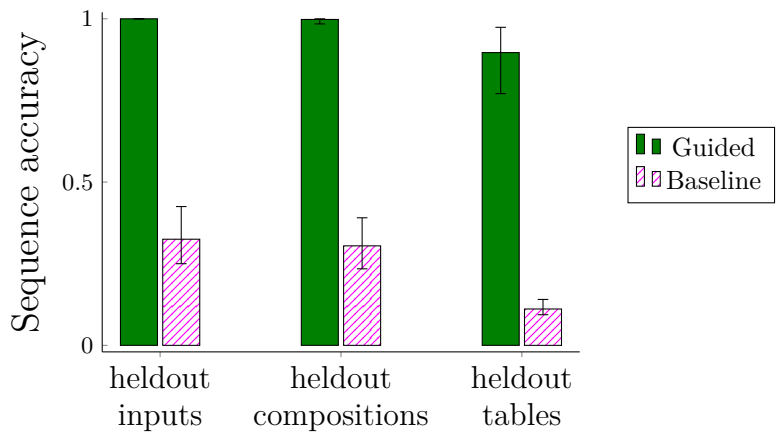
\begin{figure}[!ht]
    \centering
\begin{tikzpicture}
\begin{axis}[
    ybar,
    axis x line*=bottom,
    axis y line*=left,
    symbolic x coords={heldout inputs, heldout compositions, heldout tables},
    width=0.65\linewidth, height=0.5\linewidth,
    anchor=above north,
    ymin = -0.02,
    ymax=1.02,
    tickwidth=2pt,
    bar width=14pt,
    enlarge x limits = 0.16,
    xtick={heldout inputs, heldout compositions, heldout tables},
    xticklabels={heldout\\inputs, heldout\\compositions, heldout\\tables},
    ytick={0.0, 0.5, 1},
    xticklabel style={align=center, anchor=north, font=\large},
    ylabel= Sequence accuracy,
    ylabel style={font=\Large},
    legend style={at={(1.10,0.55)}, font=\normalsize, align=center, anchor=west}
    ]
        \addplot [style={fill=green_mtplotlib},error bars/.cd, y dir=both,y explicit]  
        table [x=compName, y=seqacc,y error plus expr=\thisrow{max}-\thisrow{seqacc}, y error minus expr=\thisrow{seqacc} - \thisrow{min}, col sep=comma,] {data/pre_rnn_results_learned_.csv};		
        \addplot [pattern=north east lines, pattern color=magenta_mtplotlib, error bars/.cd, y dir=both,y explicit] 
        table [x=compName, y=seqacc, y error plus expr=\thisrow{max}-\thisrow{seqacc}, y error minus expr=\thisrow{seqacc} - \thisrow{min}, col sep=comma] {data/pre_rnn_results_baseline_.csv};
        \legend{Guided, Baseline}
\end{axis}
\end{tikzpicture}
\caption{Average accuracies on all different test sets for chosen configurations for the lookup-table task. 
    The guided models (green) strongly outperform the baseline models (magenta) in all different test conditions and exhibit a low variance across runs.
The error bars indicate the performance of the worst and best performing model on an individual testset.}\label{fig:results_all_accuracies}
\end{figure}

For both the baseline and the guided models we pick the configuration that has the best performance across the heldout data.
We find the best baseline configuration to be a model with an embedding size of 128 and a hidden layer size of 512, whereas the best guided configuration has 16 and 512 nodes for embeddings and hidden layers, respectively.
To evaluate the chosen configurations, we generate 4 new instances of training and testing data, generated with similar criteria, but with different instances.
For each configuration and sample we train two models, 
resulting in 8 different runs for both baseline and guided models.
The average accuracies can be found in Figure \ref{fig:results_all_accuracies}.

As expected, all baseline models fail to capture a compositional strategy and perform poorly across all test sets. 
All guided models, however, achieve a near perfect generalisation accuracy to the simplest case of generalisation 
and also generalise well to compositions that are not seen at all during training (\textit{heldout compositions}).
For the most difficult case of compositionality that requires using tables compositionally that are only seen atomically (\textit{heldout tables}) the generalisation accuracy goes slightly down, and not all models converge to a perfect performance.

Overall, these results provide a clear demonstration that attentive guidance very effectively directs models to more compositional solutions. 
Later, in the analysis section of this paper, we try to push the generalisation skills of the model even further, and provide a more in-depth analysis of the guided models.


\subsection{Symbol rewriting}

We perform again a modest grid search over embedding sizes \{32, 64\} and hidden layer sizes \{64, 128, 256\}.
For the baseline model, we find that the configuration of 64x64 performs best, whereas the guided models require more parameters (32x256).\footnote{Given the high variance in the baseline performance, we ran our experiments also with the baseline model that has the same dimensions as the best guided model, but we did not find strong differences.} As for attention mechanisms, both the baseline and the guided models using the \textit{mlp} alignment and \textit{pre-rnn} perform best.
Following \cite{weber2018fine}, we train all configurations 50 times and use the validation performance as stopping criterion.

\begin{figure}[ht]
    \centering
\begin{tikzpicture}
\begin{axis}[
    ybar,
    axis x line*=bottom,
    axis y line*=left,
    symbolic x coords={Standard, Repeat, Short, Long},
    width=0.7\linewidth, height=.5\linewidth,
    anchor=above north,
    ymax=1.02,
    tickwidth=2pt,
    bar width=14pt,
    enlarge x limits = 0.16,
    xtick=data,
    ymin = -0.02,
    ytick={0.0, 0.5, 1},
    xticklabel style={font=\large},
    ylabel=sequence accuracy,
    yticklabel pos=right,
    ylabel style = {font=\Large}, 
    legend style={at={(1.10,0.55)}, font=\normalsize, align=center, anchor=west}
    ]
        \addplot [style={fill=green_mtplotlib},error bars/.cd, y dir=both,y explicit]  
        table [x=compName, y=seqacc,y error plus expr=\thisrow{max}-\thisrow{seqacc}, y error minus expr=\thisrow{seqacc} - \thisrow{min}, col sep=comma,] {data/sr_guided.csv};		
        \addplot [pattern=north east lines, pattern color=magenta_mtplotlib, error bars/.cd, y dir=both,y explicit] 
        table [x=compName, y=seqacc, y error plus expr=\thisrow{max}-\thisrow{seqacc}, y error minus expr=\thisrow{seqacc} - \thisrow{min}, col sep=comma] {data/sr_baseline64.csv};
        \legend{Guided, Baseline}
\end{axis}
\end{tikzpicture}
\caption{Average accuracies on all different test sets for chosen configurations for the symbol rewriting task. 
    The guided models (green) outperform the baseline models (magenta) in all different test conditions and exhibit a low variance across runs.
The error bars indicate the performance of the worst and best performing model on an individual testset.}\label{fig:sr_results}
\end{figure}
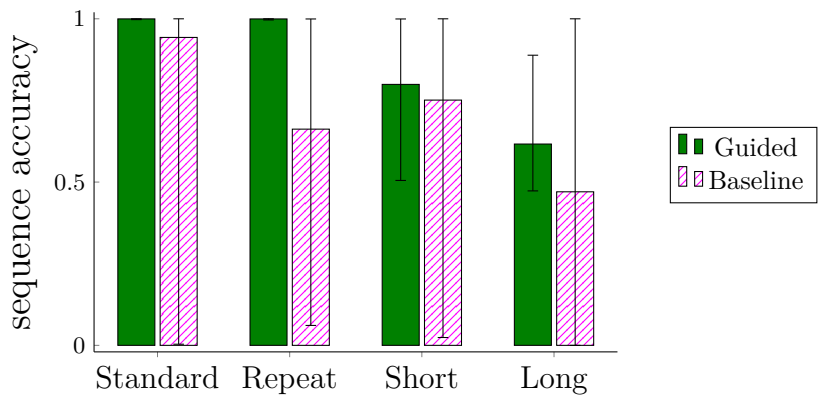

\begin{figure}[!ht]
\begin{subfigure}[b]{0.245\linewidth}
    \centering
    \includegraphics[width=\textwidth]{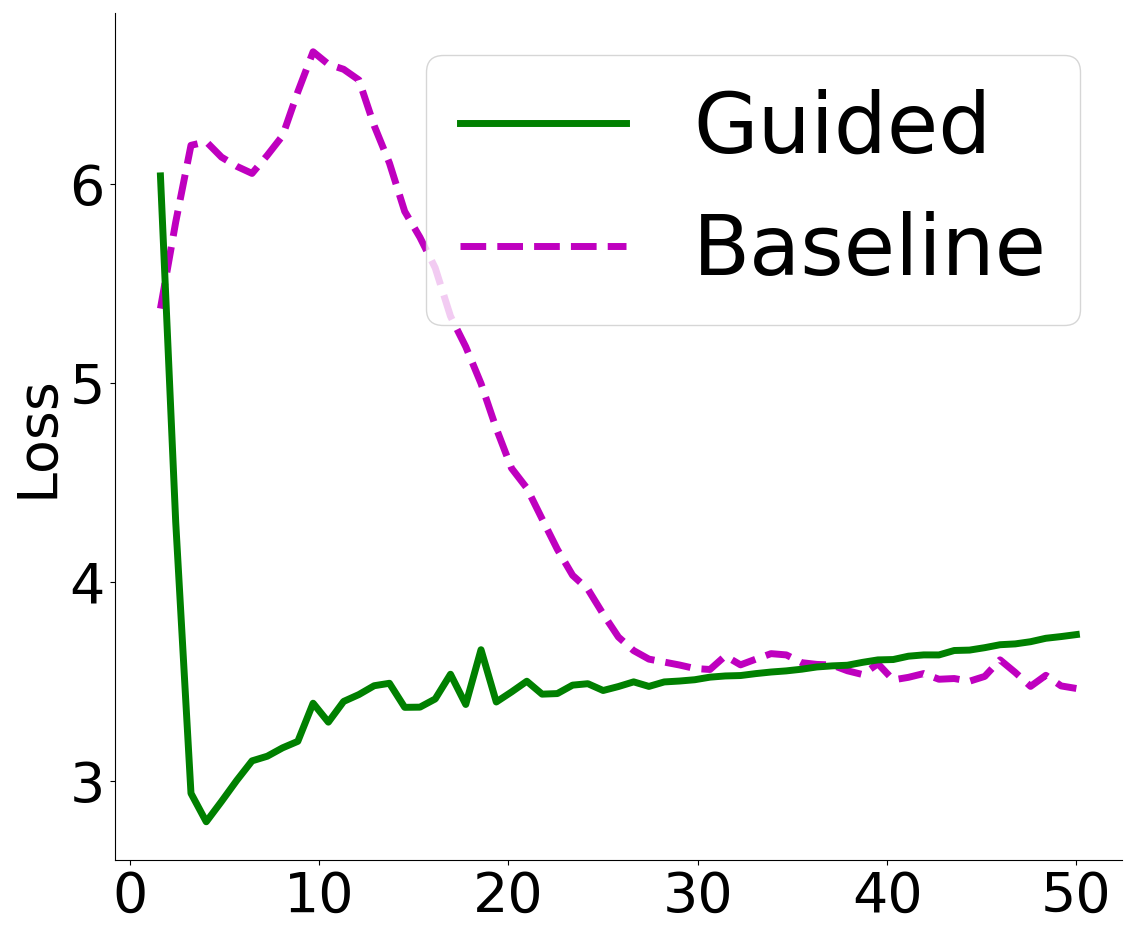}
    {{\small Standard}}    
\end{subfigure}
\begin{subfigure}[b]{0.245\linewidth}  
    \centering
    \includegraphics[width=\textwidth]{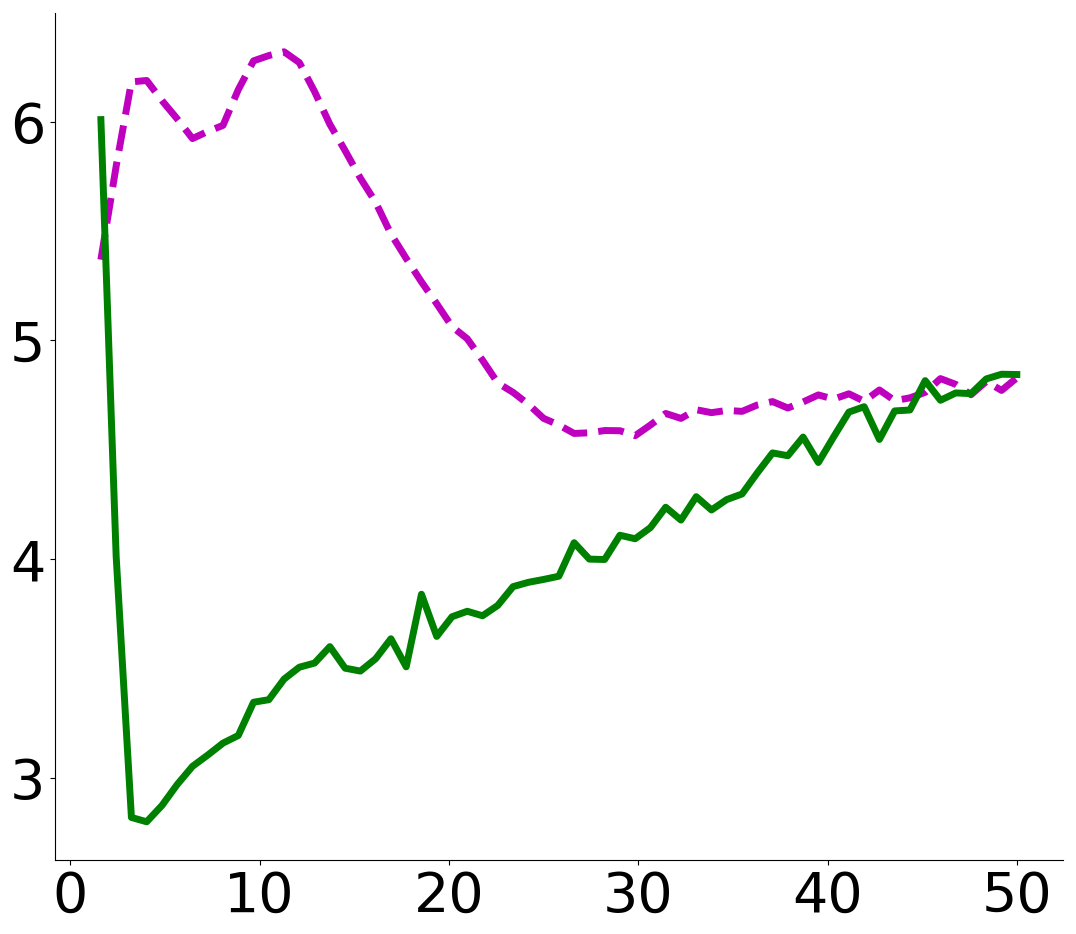}
    {{\small Repeat}}    
\end{subfigure}
\begin{subfigure}[b]{0.245\linewidth}   
    \centering
    \includegraphics[width=\textwidth]{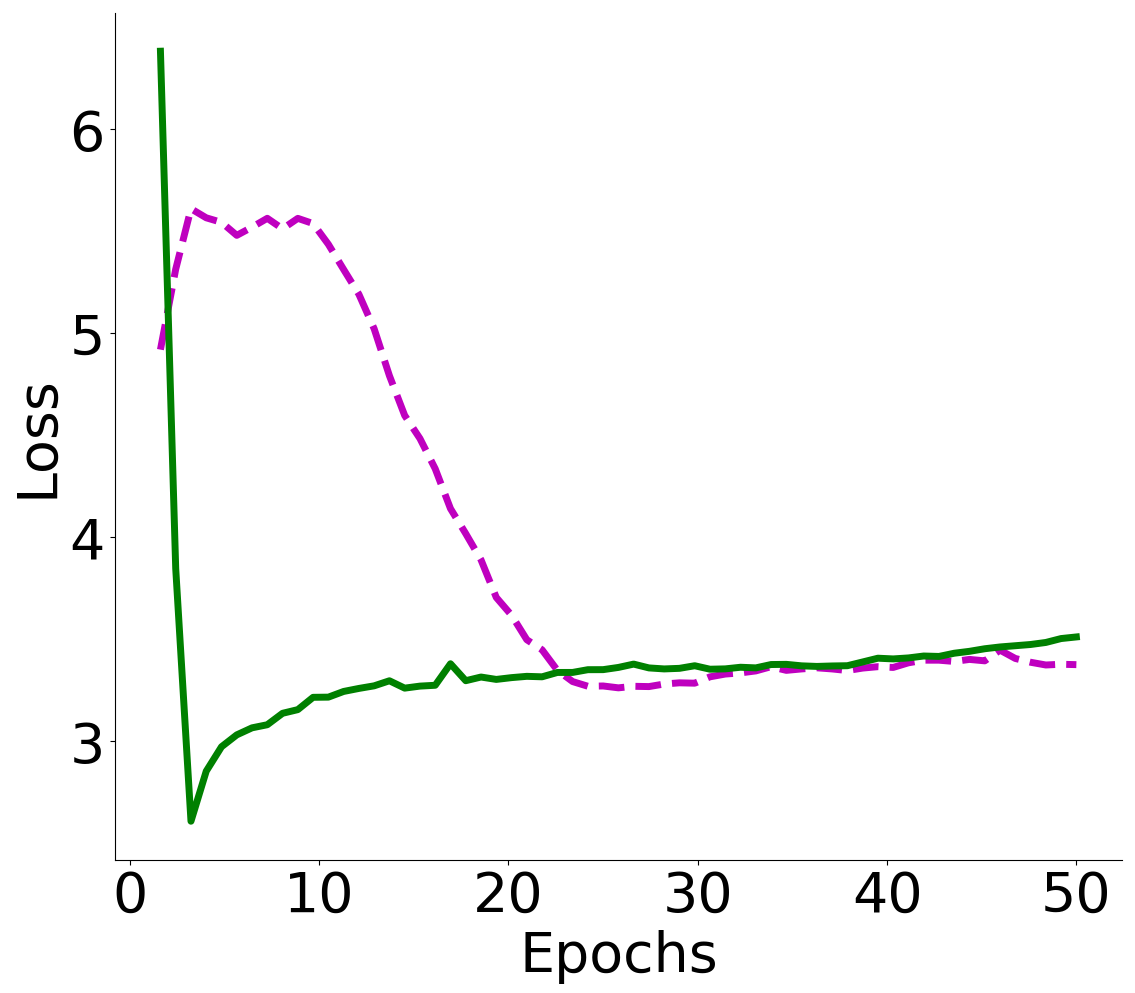}
    {{\small Short}}    
\end{subfigure}
\hfill
\begin{subfigure}[b]{0.245\linewidth}   
    \centering
    \includegraphics[width=\textwidth]{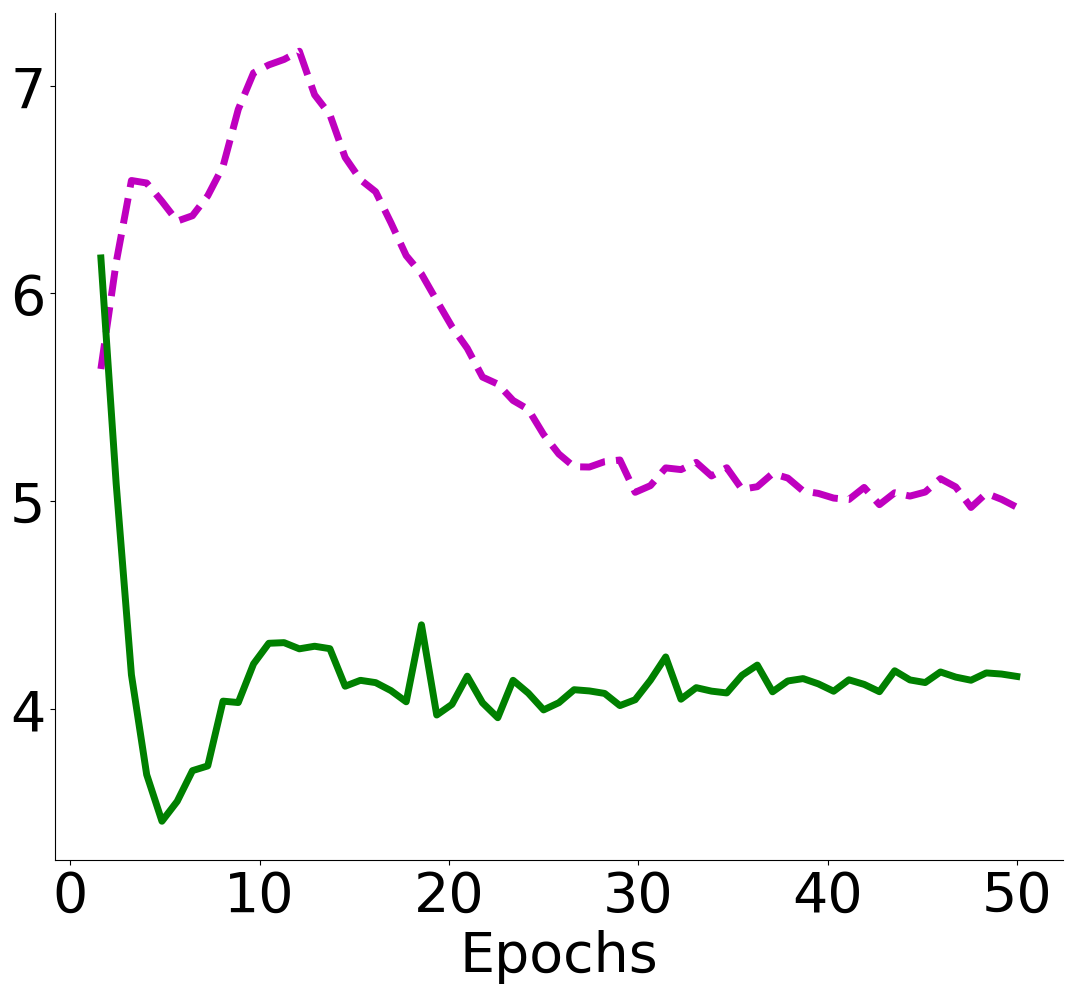}
    {{\small Long}}    
\end{subfigure}
\caption{Loss development for chosen configurations on all four test sets of the symbol rewriting task.
    Both the guided (green) and the baseline (magenta) model exhibit overfitting on the symbol rewriting task, but the guided models take much less time to converge.}\label{fig:overfitting}
\end{figure}

In Figure~\ref{fig:sr_results}, we show the average performance of the guided models (green) and the baseline (magenta), using the error bars to indicate the score of the worst and best performing model, respectively.
Although our scores are somewhat higher than those found by \cite{weber2018fine}, they do confirm that no baseline model can find a solution covering all test set distributions.
Overall, the guided models exhibit both a considerably larger accuracy accross test sets, and show significantly less variance among different runs.
Additionally, the guided models converge much quicker than the baseline models (see Figure \ref{fig:overfitting}).
In conclusion, using AG leads also to an improvement in the symbol writing task, but does not completely solve it.

\section{Analysis}
\label{sec:analysis}

With our experiment we showed that AG consistently improves the compositional skills of seq2seq models.
In particular, we observed that AG helps to decrease the variance between different initialisations and strongly countereffects overfitting.
We now provide a brief analysis of the learned models and assess the influence of the different components of AG.


\subsection{Attention Patterns}

\begin{figure}[]
    \centering
\begin{subfigure}{0.39\linewidth}
    \includegraphics[width=\linewidth, trim={35mm, 20mm, 20mm, 20mm}, clip]{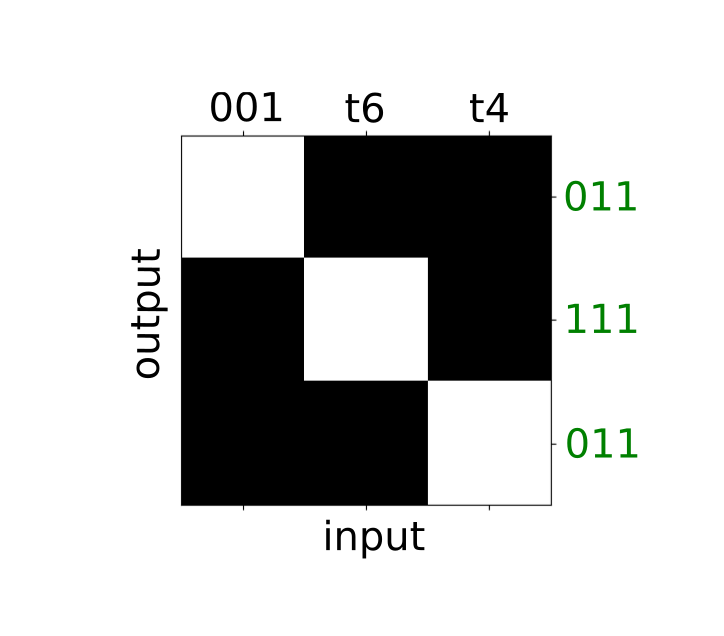}
    \vspace{-3mm}
\end{subfigure}
\hspace{3mm}
\begin{subfigure}{0.39\linewidth}
    \includegraphics[width=\linewidth, trim={10mm, 00mm, 00mm, 10mm}, clip]{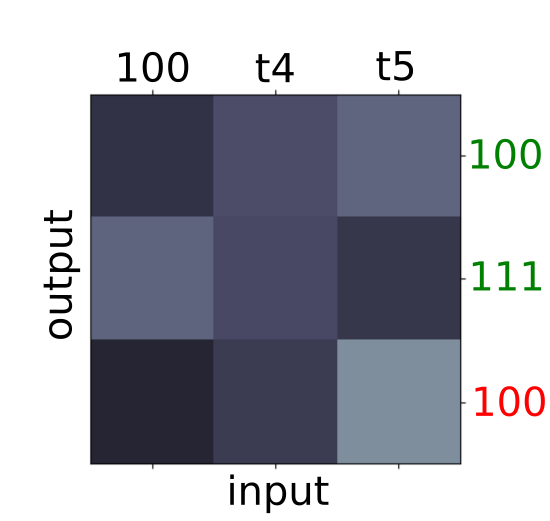}
\end{subfigure}
    \caption{Attention plots for a heldout composition processed by a guided model (left) and a heldout input processed by a baseline model (right). The x-axis represents the input, the y-axis the output, the plot color represents the strength of the attention weights (black=0, white=1). E.g., we see that while generating the first output (represented in the first row), the attention of the guided model is focused on the first symbol of the input, while the baseline attention is diffused.}\label{fig:attn_plots_lookup}
\end{figure}

\begin{figure}[!ht]
\centering
\includegraphics[width=0.45\linewidth, trim={40mm, 22mm, 0mm, 22mm}, clip]{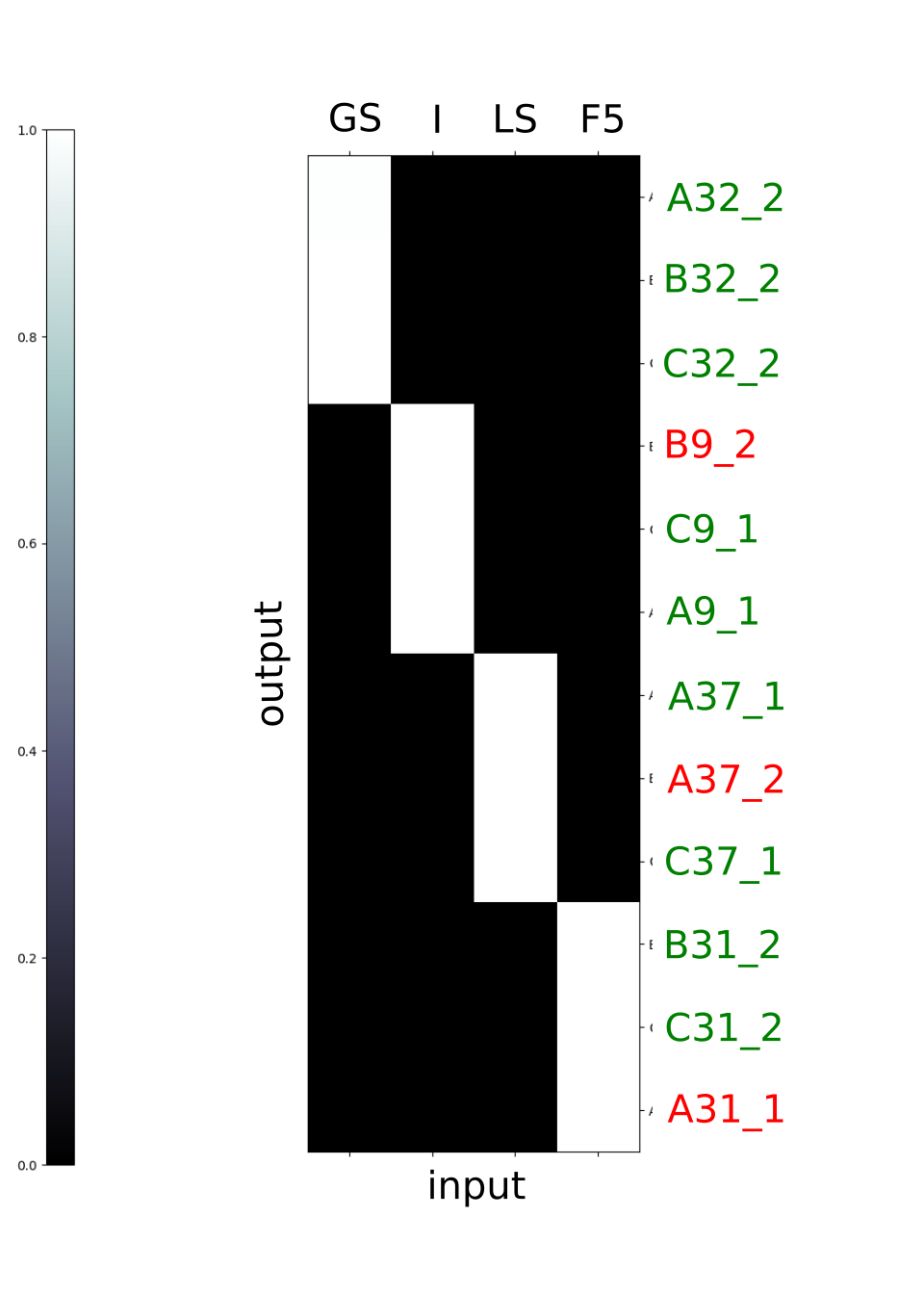}
\includegraphics[width=0.45\linewidth, trim={40mm, 22mm, 0mm, 22mm}, clip]{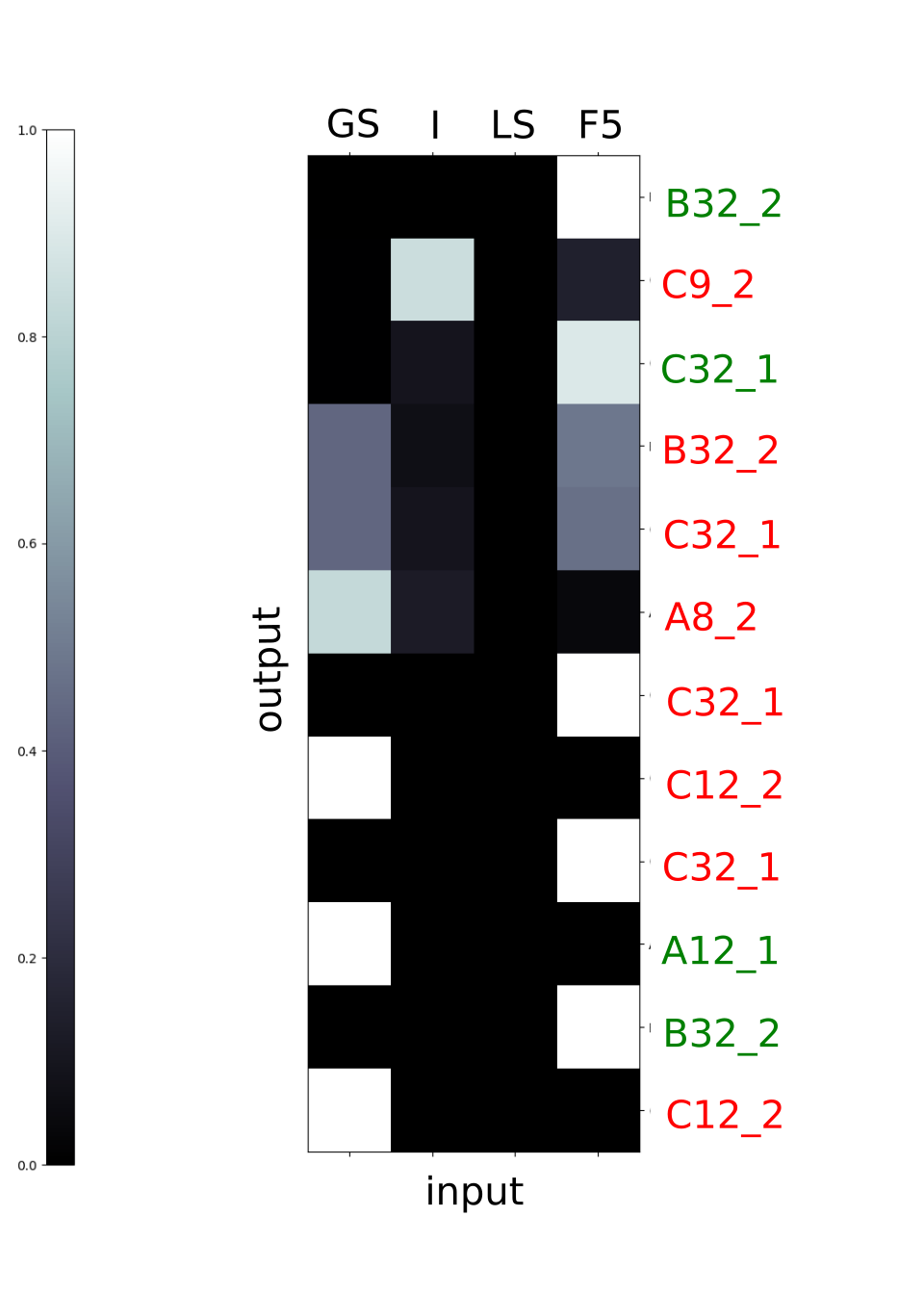}
\caption{The attention patterns for a short composition in the symbol rewriting task. The guided model (left) focuses on the input symbols that are relevant for the output symbol it is currently generating but the baseline model (right) does not.}\label{fig:attn_plots_sg}
\end{figure}

The attention plots demonstrate that the guided models learned to generate an (almost) perfect attention pattern, which focuses on the right inputs at the right time.
The baseline models, on the other hand, do not use their attention mechanism in a targetted way, as indicated by their diffused attention. 

\subsection{Sparse attentions}
We hypothesise that the effect of AG could be due to the sparsity of the learned attention vectors, and investigate to what extent merely enforcing this sparsity can match the effect of training with an AG target.
We repeat experiments with a vanilla seq2seq model for which we ensure peakiness of the attention using Gumble softmax \cite{jang2016categorical}, but we did not find similar improvements.

\subsection{Oracle Guidance}

\begin{figure}
    \begin{subfigure}{.46\linewidth}
        \vspace{1mm}
        \includegraphics[width=0.95\linewidth, trim={53mm, 5mm, 17mm, 5mm}, clip]{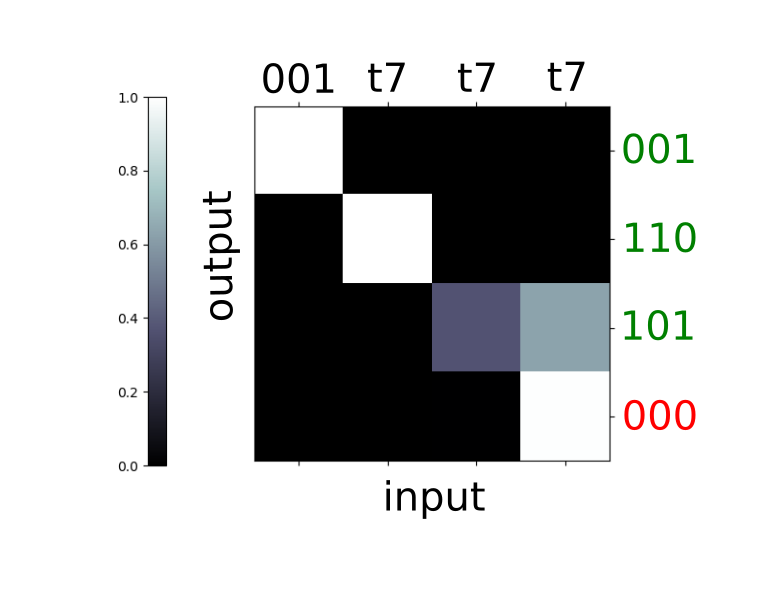}
    \end{subfigure}
    \begin{subfigure}{.53\linewidth}
        \includegraphics[width=0.95\linewidth]{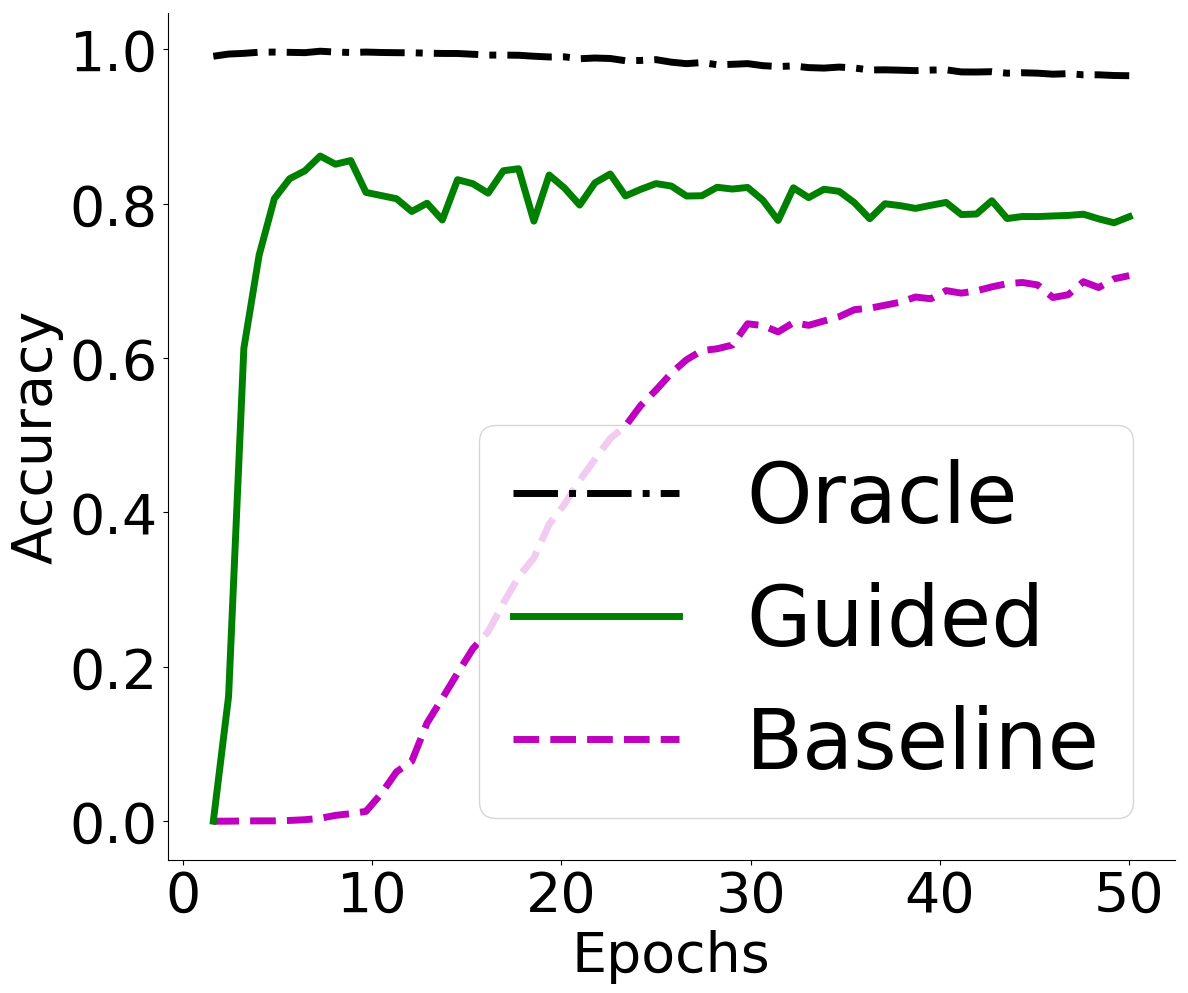}
    \end{subfigure}
    \caption{An attention pattern for longer lookup table compositions (left) and accuracies on symbol rewriting for baseline, guided and oracle attention (right).}\label{fig:longer}
\end{figure}

When probing the guided lookup table models to generalise to longer sequences, we observe that their attention becomes more diffuse (see Figure~\ref{fig:longer}).
This failure in producing the right attention patterns matches the performance of the models on longer compositions, which rapidly drops when the compositions get longer.
To test to what extent this inability stems from the model's difficulty to generalise the attention patterns to longer sequences, we investigate if we can push models to find solutions for the task when instead of learning the attention patterns, we use \textit{oracle guidance}.

Our experiments with oracle guidance show that when the right attention vector is provided to a model a perfect performance is reached almost always for both explored tasks.
The difference between train and test performance are flattened even in very extreme conditions, such as the very long lookup table sequences.
We conjecture that this finding begs into question whether the current architecture optimally facilitates AG, and we propose that perhaps it would be more suitable to use two distinct models to learn the target outputs and the correct attention patterns.


\section{Conclusion}
In this paper, we focus on directing recurrent neural networks to find more compositional solutions, focusing in particular on seq2seq models equipped with an attention mechanism.
To do so, we use a technique we call \textit{attentive guidance}, in which we guide the attention mechanism of a seq2seq model to attend to the individual components of the input sequence in a compositional way by providing an extra loss term at training time.
We show that attentive guidance can successfully direct models to find solutions for two tasks that were specifically designed to evaluate compositional abilities while leaving out other possibly confounding signals.

A strong limitation of AG is the need of an additional supervision signal that may often not be available -- especially when dealing with real world tasks.
One potential direction for future work, which could potentially be interesting from a cognitive or linguistic perspective, is to investigate whether metadata available for some natural corpora (consider, e.g., prosody information), could be injected in models via the attention mechanism.
Another promising research path would explore methods that automatically learn the pattern from data, or incorporate the positive effects of attentive guidance in the architecture in another way, one could explore methods that would automatically learn the pattern from data, and as such create a compositional solution not by explicitly adding it, but by simply modulating information flow.

In summary, with these results we confirm that (i) provided the right objective function, seq2seq models trained with gradient descent can find compositional solutions without being enhanced with additional component that explicitly models compositionality; (ii) one of the obstacles that prevents vanilla seq2seq models to do so is their difficulty to distinguish salient patterns from spurious ones.
Such findings provide strong support for keep focusing on seq2seq models to tackle compositionality, with the big advantage of maintaining rather simple and easy to train models. 
\bibliographystyle{splncs}
\bibliography{refs}

\end{document}